\documentclass[11pt,a4paper]{article}
\usepackage[hyperref]{eacl2021}
\usepackage{times}
\usepackage{latexsym}
\usepackage{amsmath}

\usepackage{todonotes}
\usepackage{multirow}
\usepackage{colortbl}
\usepackage{ulem}
\usepackage{caption}
\usepackage{subcaption}
\usepackage{amsfonts}
\usepackage{arydshln}
\usepackage{algorithm}
\usepackage{algpseudocode}

\usepackage{microtype}

\aclfinalcopy % Uncomment this line for the final submission
\title{Zero-Shot Cross-Lingual Dependency Parsing through Contextual Embedding Transformation}

\author{Haoran Xu \and Philipp Koehn\\
Johns Hopkins University\\
\texttt{hxu64@jhu.edu, phi@jhu.edu}\\
}
\date{}

\begin{document}
\maketitle
\begin{abstract}
Linear embedding transformation has been shown to be effective for zero-shot cross-lingual transfer tasks and achieve surprisingly promising results. However, cross-lingual embedding space mapping is usually studied in static \textit{word-level} embeddings, where a space transformation is derived by aligning representations of translation pairs that are referred from dictionaries. We move further from this line and investigate a contextual embedding alignment approach which is \textit{sense-level} and dictionary-free. To enhance the quality of the mapping, we also provide a deep view of properties of contextual embeddings, i.e., the anisotropy problem and its solution. Experiments on zero-shot dependency parsing through the concept-shared space built by our embedding transformation substantially outperform state-of-the-art methods using multilingual embeddings.

\end{abstract}

\section{Introduction}
Cross-lingual embedding space alignment \citep{mikolov2013exploiting,artetxe-etal-2016-learning,xing-etal-2015-normalized,lample2018word} recently has been attracted a lot of attention because cross-lingual model transfer is effectively facilitated by shared semantic spaces in NLP tasks, e.g., named entity recognition \citep{xie-etal-2018-neural}, part-of-speech tagging \citep{hsu-etal-2019-zero}, and dependency parsing \citep{schuster-etal-2019-cross}, where dependency paring is scoped out in this paper. Compared with the delexicalized parsers \citep{mcdonald-etal-2011-multi}, multilingual word embeddings have been demonstrated to significantly improve the performance of zero-shot dependency parsing by bridging the lexical feature gap \citep{guo-etal-2015-cross}.

\begin{figure*}[h]
    \centering
    % \hbox{\hspace{1.5em}\includegraphics[width=8cm]{figure/model.png}}
    \includegraphics[width=15.5cm]{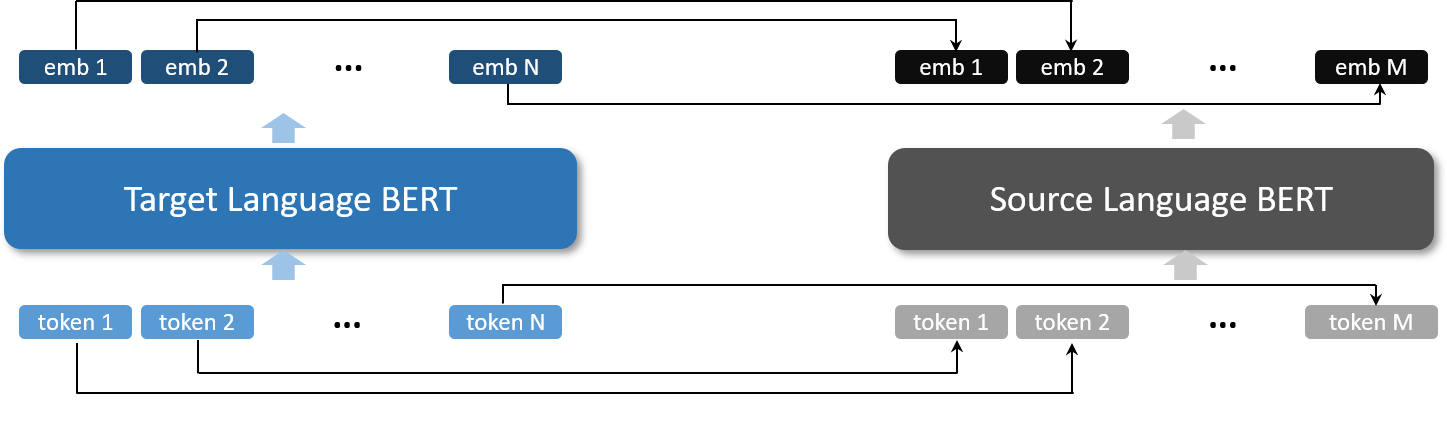}
    \caption{The target tokens (left, blue) and the source tokens (right, black) are aligned by \textit{Fast Align}, so their contextual embeddings can be aligned as well.}
    \label{fig:link}
\end{figure*}

With the remarkable development of monolingual contextual pre-trained models \citep{peters-etal-2018-deep,devlin-etal-2019-bert,radford2019language}, which dramatically outperform static word embeddings \citep{mikolov2013efficient,pennington2014glove,bojanowski-etal-2017-enriching} in broad NLP applications, increasing number of researchers have started focusing on contextual representation alignment for cross-lingual dependency parsing \citep{schuster-etal-2019-cross,wang2019cross}. Moreover, with the appearance of multilingual pre-trained models, such as Multilingual BERT (mBERT) \citep{devlin-etal-2019-bert}, zero-shot dependency parsing becomes easier by utilizing the large vocabulary of the multilingual models \citep{kondratyuk201975}.

Our approach is most similar to \citet{schuster-etal-2019-cross}, which maps a target language space into a source language space through a linear transformation to realize zero-shot transfer in dependency parsing. Typically, a transformation is usually derived by word-level embedding alignment, while we explore a sense-level embedding alignment method to map bilingual spaces more precisely, where sense-level representations are split from multi-sense word-level embeddings. Furthermore, our mapping approach is dictionary-free which utilizes the silver token pairs from parallel corpora and eliminates the necessity of gold dictionaries. The experimental results of zero-shot dependency parsing demonstrate that two parser evaluation scores (UAS and LAS) of sense-level mapping are always better than of word-level one. Moreover, we also notice the anisotropy problem \citep{ethayarajh-2019-contextual} (defined in Section \ref{subsec:ansiotropy}) in contextual embeddings, which potentially deteriorate the performance of the zero-shot transfer task. We significantly mitigate this drawback by leveraging a prepossessing step, \textbf{i}terative \textbf{n}ormalization (IN) \citep{zhang-etal-2019-girls}, which is originally used for improving the performance of static embedding mapping on the bilingual dictionary induction task.
%However, instead of leveraging mean contextual vector from pre-trained ELMo \citep{peters-etal-2018-deep} model to align bilingual spaces with a static dictionary, our mapping exploits parallel corpus and pre-trained BERTs.

Zero-shot dependency parsing experiments are conducted on \textit{Universal Dependencies treebank v2.6} \cite{11234/1-3226}, which shows that our results obtain a substantial gain compared with state-of-the-art methods using multilingual fastText and mBERT \footnote{Code is available at: \url{https://github.com/fe1ixxu/ZeroShot-CrossLing-Parsing}.}.

\section{Linear Cross-lingual Space Alignment}
Let denote $X\in\mathbb{R}^{d\times N}$ as the word embedding matrix for a target language \footnote{Different from usual settings, we use $x$-related symbols for target data and $y$-related ones for source data.}, and $Y$ as the word embedding matrix for a source language. For each column of the target embedding matrix $x_i\in \mathbb{R}^d$, it has one source embedding vector $y_i\in \mathbb{R}^d$ corresponding to a source word translated from the target word $i$. We aim to derive a linear transformation matrix $\hat{W}$ used for mapping from the target language space to the source language space. This can be learned by minimizing the Frobenius norm: 
\begin{equation}\
    \label{eq:min}
    \hat{W} = \mathop{\arg\min}_{W\in \mathbb{R}^{d\times d}}\|WX-Y\|_F
\end{equation}
Furthermore, \citet{xing-etal-2015-normalized} show that the quality of space alignment is successfully improved with the orthogonal restriction, i.e, $W^TW=I$. Thus, the problem can be solved by Procrustes approach \citep{schonemann1966generalized}:
\begin{equation}\
    \label{eq:procruste}
    \begin{split}
        \hat{W} = &\mathop{\arg\min}_{W\in \mathbb{O}^{d\times d}}\|WX-Y\|_F = UV^T\\
        &\ \text{s.t.} \ \ U\Sigma V^T = \text{svd}(YX^T)
    \end{split}
\end{equation}
where $\mathbb{O}^{d\times d}$ is the set of orthogonal matrices.

\section{Method}
\subsection{Contextual Embedding Transformation}
An unsupervised bidirectional word alignment algorithm based on \textit{IBM Model 2} \citep{brown-etal-1993-mathematics}, \textit{Fast Align} \citep{dyer-etal-2013-simple}, is first applied to a parallel corpus to derive silver aligned token pairs. We then respectively feed the parallel corpus to the BERTs of the target and the source languages and extract the outputs as contextual embeddings. As shown in Figure \ref{fig:link}, \textit{Fast Align} bridges ``links" between silver token pairs, and between the embeddings of the token pairs as well. Thus, for each target type, a collection of its contextual embeddings can be obtained, as well as a collection of contextual embeddings of its aligned source tokens.  Vectors are normalized to satisfy the orthogonal condition.

Motivated by the assumption that multiple senses of a type can construct multiple distinct clusters in its collection \citep{schuster-etal-2019-cross}, we derive several sense-level (cluster-level) embeddings for a type by averaging vectors in each cluster. This splits the representations of multi-sense words and helps the anchor-driven space mapping in a finer resolution. To find clusters, we utilize $k$-means to cluster contextual embeddings in the vector collection of each type, and adaptively find the optimal $k$ by an elbow-based method \citet{satopaa2011finding}. Contextual vectors are only clustered in the target side to obtain sense-level embeddings, while the aligned sense-level embeddings in the source side can also be simultaneously derived because embeddings have been already ``linked" by \textit{Fast Align}. We next build a sense embedding matrix $X_s$ for the target language by putting the sense-level embeddings in each column, and meanwhile construct a column-wise aligned sense embedding matrix $Y_s$ in the source side. Finally, we obtain the optimal linear mapping $\hat{W}$ from $X_s$ to $Y_s$ by Equation \ref{eq:procruste}. Pseudo code of transformation method is in Appendix \ref{appendix:algo}.

% finding from \citet{schuster-etal-2019-cross} that the collection of contextual embeddings for a token can form a cluster in the semantic space, and clusters corresponding to different tokens are well separated, we extend this hypothesis to the sense-level. Namely, well separated clusters are only composed of the contextual embeddings which represent the same meaning. 
% \footnote{We also tried other cluster algorithms, e.g., agglomerative clustering, gaussian mixtures. Our preliminary results show that $k$-means performs best.}

\begin{figure}
     \centering
     \begin{subfigure}[b]{0.23\textwidth}
         \centering
         \includegraphics[width=\textwidth]{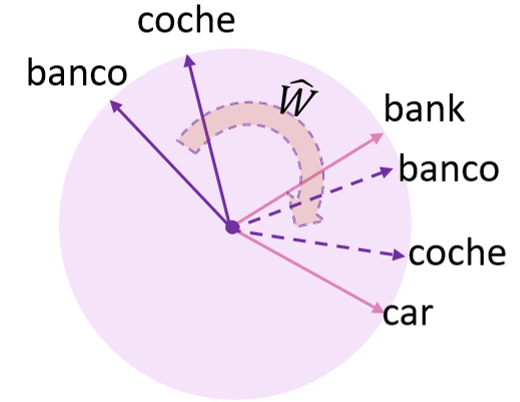}
         \caption{}
         \label{fig:beforeIN}
     \end{subfigure}
     \hfill
     \begin{subfigure}[b]{0.23\textwidth}
         \centering
         \includegraphics[width=\textwidth]{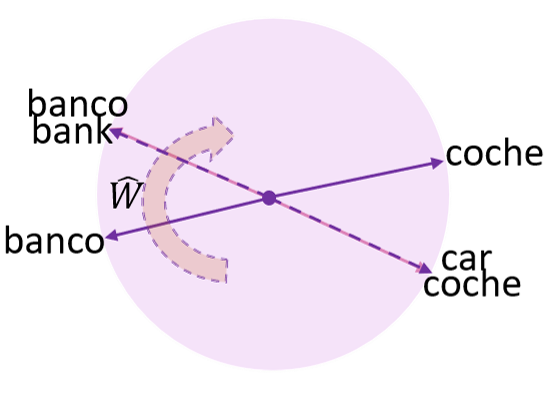}
         \caption{}
         \label{fig:afterIN}
     \end{subfigure}
     \caption{(a) Spanish vectors (purple arrows) cannot well fit to English vectors (pink arrows) by a linear transformation because they gather in different degrees of cones (different angles between vectors), where dash lines are mapped vectors. (b) After iterative normalization, Spanish and English vectors are uniformly distributed (same angles between vectors). They can be perfectly fit after mapping now. }
     \label{fig:anisotropy}
\end{figure}

\subsection{Anisotropy in Embedding Spaces}
\label{subsec:ansiotropy}
Our findings show that contextual embeddings always hold anisotropic property, i.e., they are not uniformly distributed in the space and gather toward a narrow range of orientations. Importantly, degrees of anisotropy across languages are various, which undermines the quality of cross-lingual mappings. A toy example of how the anisotropy affect mappings is illustrated in Figure \ref{fig:beforeIN}. One metric for anisotropy is to calculate the average cosine similarity distance of randomly selected vectors. The higher the distance is, the narrower directions vectors point to. Note that the distance for an isotropic space is 0. To mitigate this problem, we introduce iterative normalization. For each token $i$, the embedding vector $x_i$ is forced to be zero-mean firstly in each iteration:
\begin{equation}
    x_i = x_i - \frac{1}{N}\sum_{i=1}^Nx_i
\end{equation}
and then normalize it to a fixed length:
\begin{equation}
    x_i = \frac{x_i}{\|x_i\|_2}
\end{equation}
The two steps are repeated until convergence. $N$ is the total number of embeddings. The iterative preprocessing enforces the space to be uniformally distributed, and relative angles between vectors across languages to be more similar (Figure \ref{fig:afterIN}).

\begin{figure}[t]
    \centering
    % \hbox{\hspace{1.5em}\includegraphics[width=8cm]{figure/model.png}}
    \includegraphics[width=7.5cm]{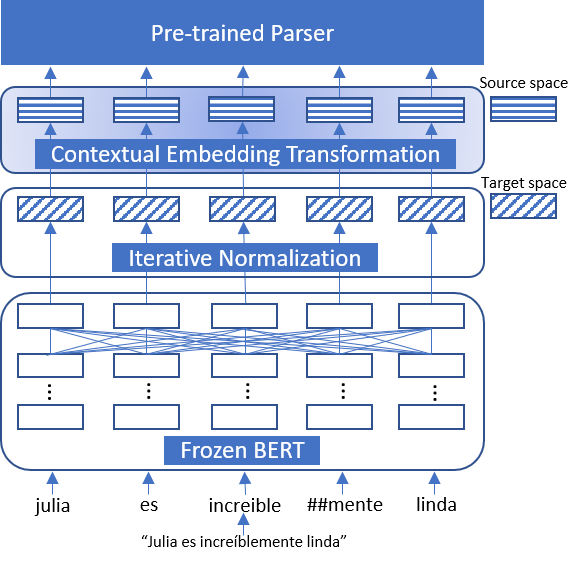}
    \caption{The workflow of how zero-shot transfer processes in our model architecture.}
    \label{fig:model}
\end{figure}

\subsection{Zero-shot Transfer}
A parser is first trained on a source language treebank, where outputs of a frozen BERT are used as embeddings. To apply the pre-trained parser to the target languages, we first replace the source BERT with the target BERT. Then, iterative normalization is operated to enforce contextual embeddings in a near-isotropic space. At last, we map the embeddings to the source language space. Specifically, for each target token $i$, its contextual representation $x_i$ is mapped by $\hat{W}x_i$. The processing of zero-shot dependency parsing is visualized in Figure \ref{fig:model}. Note that the space of pre-trained model has already fit to be near-isotropic by utilizing iterative normalization during training.

\begin{table*}[ht]
\begin{center}

\begin{scriptsize}
  \begin{tabular}{c|cc|cccccc:cc:cc:cc}
 
    lang (\textit{treebank}) & \multicolumn{2}{c}{\textbf{en} (\textit{ewt})} & \multicolumn{2}{c}{\textbf{es} (\textit{gsd})} & \multicolumn{2}{c}{\textbf{pt} (\textit{gsd})} & \multicolumn{2}{c}{\textbf{ro} (\textit{rrt})} & \multicolumn{2}{c}{\textbf{pl} (\textit{lfg})} & \multicolumn{2}{c}{\textbf{fi} (\textit{tdt})} &
    \multicolumn{2}{c}{\textbf{el} (\textit{gdt})}\\ 
    \hline 
    
    & UAS & LAS & UAS & LAS  &  UAS & LAS  & UAS & LAS  & UAS & LAS  & UAS & LAS  & UAS & LAS  \\
    \hline
    aligned fastText & 88.55 & 86.36 & 73.57 & 65.13 & 72.41 & 60.69 & 60.27 & 46.79 & 75.88 & 59.48 & 62.32 & 40.78 & 71.51 & 61.46   \\
    mBERT uncased & 93.45 & 91.52 & 82.11 & 72.51 &  80.89 & 68.90 & 72.08 & 56.91 & \bf 85.27 & 69.76 & 72.76 & 49.64 & 81.72 & 68.35 \\
    mBERT cased & 93.32 & 91.34 & 82.83 & 74.08 & 80.80 & 68.68 & 70.76 & 56.04  & 83.77 & 68.01 & 71.82 & 48.84 & 78.24 & 65.92   \\
    \hline
    word-level & \multirow{2}{*}{\underline{93.70}} & \multirow{2}{*}{\underline{91.78}} & 82.43 & 73.86 & 79.77 & 67.35 & 71.13 & 57.28 & 84.58 & 69.53 & 74.65 & 51.06 & 82.29 & 69.88 \\
    sense-level &   &   & 82.55 & 73.92 & 80.34 & 67.80 & 71.46 & 57.57 & \underline{84.71} & 69.56 & 74.81 & 51.14 & 82.33 & 70.10  \\
    % \cdashline{2-3}[2pt/2pt]
    \cline{2-3}
    word-level + IN & \multirow{2}{*}{\bf 94.21} & \multirow{2}{*}{\bf 92.01} & \underline{83.70} & \underline{75.14} & \underline{81.48} & \underline{69.04} & \underline{74.65} & \underline{59.68}  & 84.32 & \underline{70.45} & \underline{75.07} & \underline{51.75}  & \underline{83.76} & \underline{71.11} \\
    sense-level + IN &   &  & \bf 83.91 & \bf 75.39 & \bf 81.99 & \bf 69.49 & \bf 74.78 & \bf 59.83 & 84.57 & \bf 70.52 & \bf 75.31 & \bf 51.99 & \bf 84.05 & \bf 71.26 \\
    \hline
  \end{tabular}
\end{scriptsize}
\end{center}
\caption{UAS and LAS of zero-shot evaluation for various languages on test files. The highest scores are bolded and the second highest scores are underlined. Language families are split by dash lines. lang = language, en = English, es = Spanish, pt = Portuguese, ro = Romanian, pl = Polish, fi = Finnish, el = Greek.}
\label{tab:results}
\end{table*}

\begin{figure*}
     \centering
     \begin{subfigure}[b]{0.45\textwidth}
         \centering
         \includegraphics[width=\textwidth]{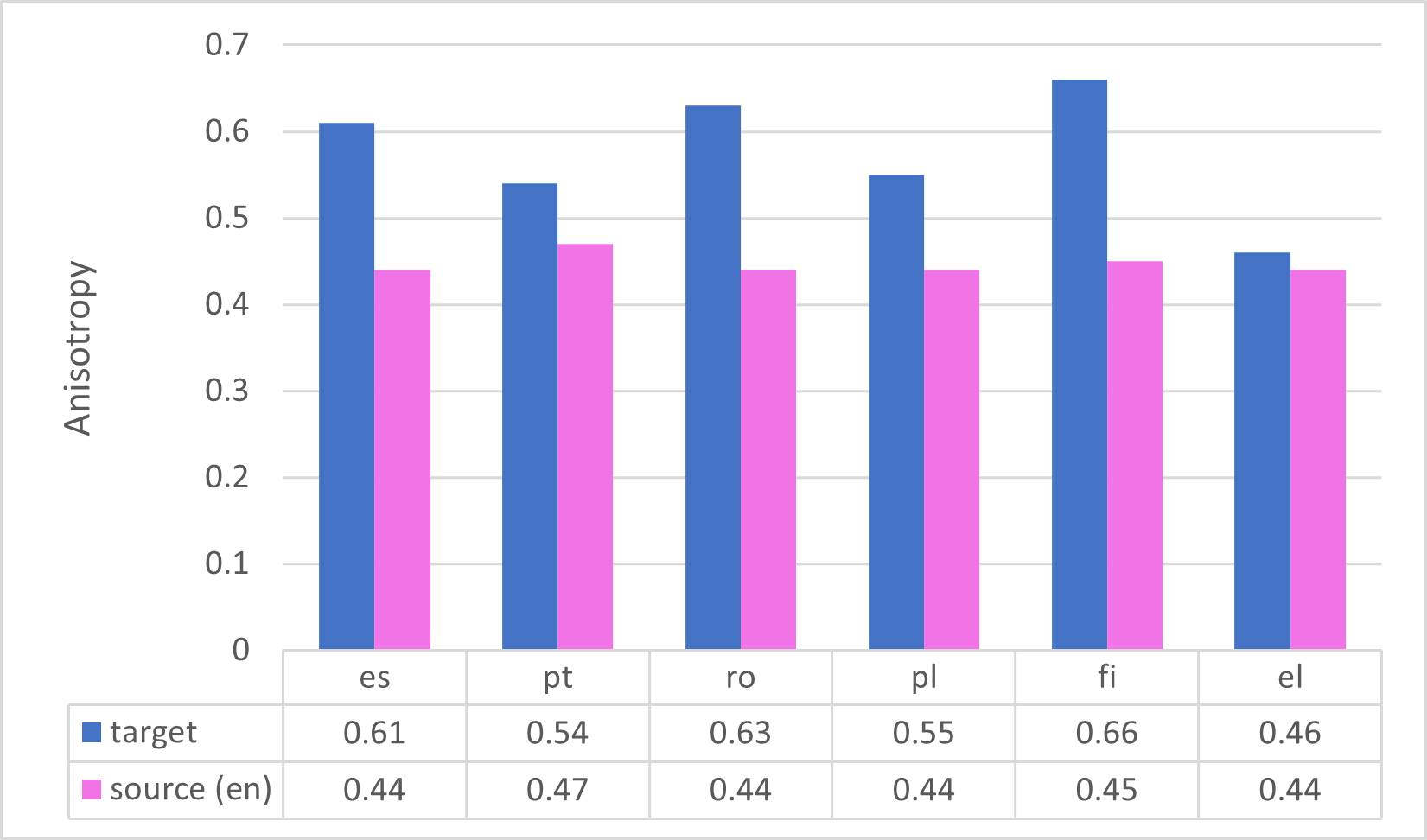}
         \caption{}
         \label{fig:horizon-chart}
     \end{subfigure}
     \hfill
     \begin{subfigure}[b]{0.45\textwidth}
         \centering
         \includegraphics[width=\textwidth]{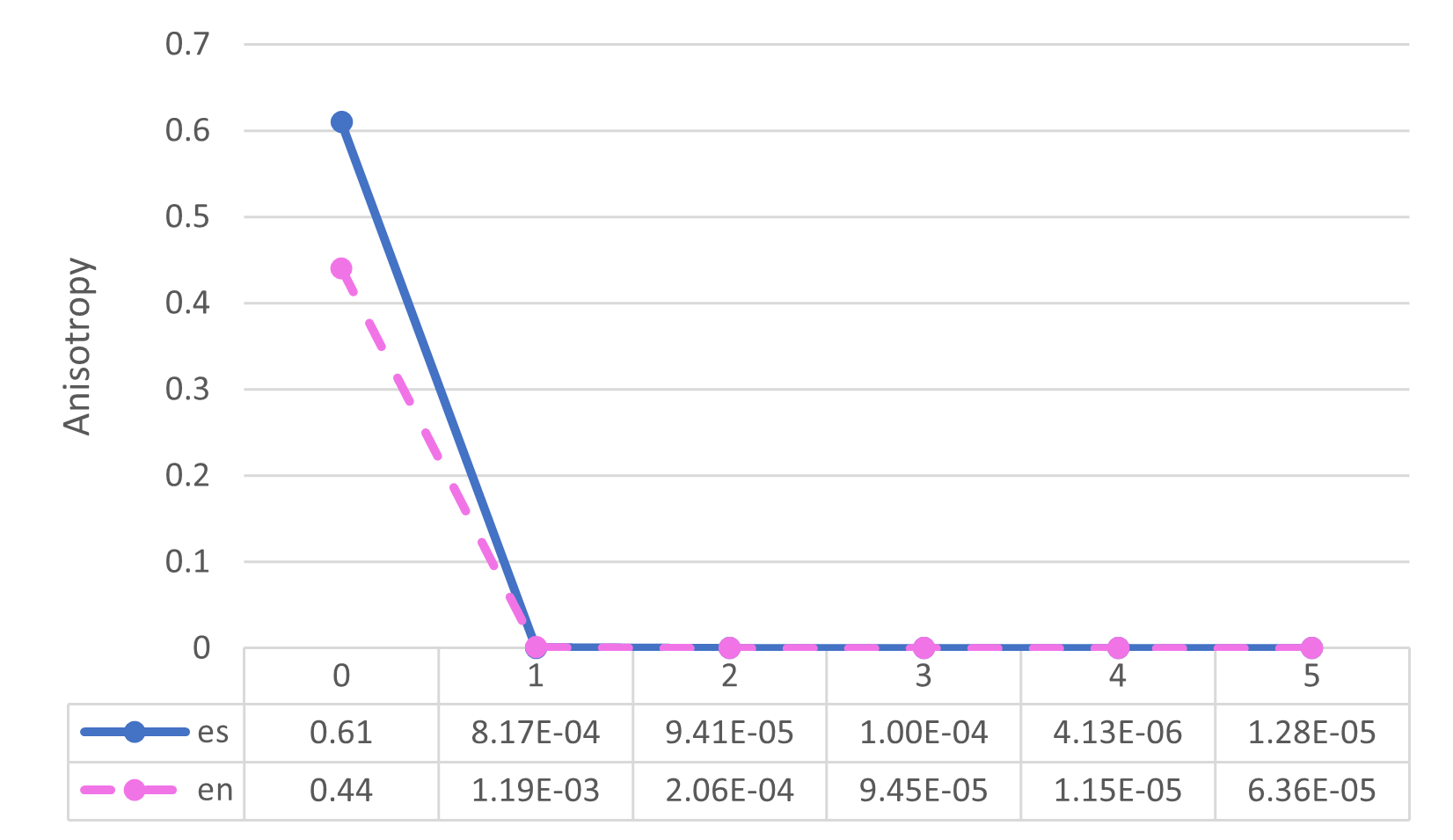}
         \caption{}
         \label{fig:IN-reduction}
     \end{subfigure}
     \caption{ (a) Discrepancy of anisotropic degrees for all tested language pairs, where scores of anisotropic degree are calculated by the mean cosine similarity between 1000 randomly selected vectors in their language spaces. (b) The isotropic degrees basically decrease to 0 at the first iteration and converge afterwards.}
     \label{fig:anisotropy-reduction}
\end{figure*}

\section{Experiment}
Our parser is the deep biaffine model from \citet{dozat2016deep} where hyperparameters are almost unchanged. The settings of all hyperparameters are listed in Appendix \ref{sec:appendix:hyper}. English is set as the source language and other languages are targets. In our experiments, we select 6 target languages from 4 language families for which we have off-the-shelf monolingual pre-trained BERT models (base-size). We train the parsing model only in the English treebank, and directly evaluate zero-shot transfer performance on the target languages.  
%Note that all BERT models are base-size. 
%except that we increase hidden size of LSTM from 400 to 500 and all of dropout rates are 0.3. , except for setting 500 for the LSTM hidden size and 0.3 for the dropout rates

\subsection{Baseline}
\paragraph{Aligned fastText:} Our first baseline is multilingual fastText aligned by the RCSLS method \citep{joulin2018loss,bojanowski-etal-2017-enriching} which is straightforwardly employed to the embedding layer for the corresponding language.

\paragraph{mBERT:} We compare our approach with both uncased and cased version of mBERT. Outputs of mBERT are directly used for the embedding layer.

\subsection{Settings}
Following the analysis that top layers of BERT contain more semantic information \cite{jawahar-etal-2019-bert}, our contextual representation are normalized mean vector of the last 4 layers of BERT. The parallel corpora used to extract contextual embeddings are obtained from \textit{ParaCrawl v6.0} \footnote{\url{www.paracrawl.eu}}. For each language pairs, we select 1M parallel sentences whose length is shorter than 150. Since some noisy alignments are produced during \textit{Fast Align}, we only take one-to-one token alignment into consideration. The dataset used for cross-lingual dependency parsing is the \textit{Universal Dependencies treebank v2.6} \footnote{\url{https://lindat.mff.cuni.cz/repository/xmlui/handle/11234/1-3226}} \cite{11234/1-3226}.%, which contains consistent treebank annotations for 85 languages.

We store up to 10K contextual vectors extracted from BERT for non-OOV tokens \footnote{We do not use the composition of subword vectors to approximately represent OOV tokens, because our preliminary results show this hurts the mapping.}. Vectors in the collection of a target type are clustered to derive sense-level embeddings only if the token occurs more than 100 times. Otherwise, the representation for the token is the basic word-level embedding, i.e., the mean vector of its vector collection. Experiments of word-level embedding alignment are also conducted to compare with sense-level results. %As for low-frequent tokens whose occurrence is lower than 20, we ignore them during the calculation of transformation matrix (Equation \ref{eq:procruste}).

%On account of the fact that the approximate representations for OOV tokens which are derived by the mean \footnote{We also tried the leftmost and rightmost subword representations for OOV tokens, but the results are similar. We assume this happens because of the lower-precise representations in the semantic space. } of subword embeddings will hurt the mapping performance, we discard OOV representation during mapping. 

\subsection{Iterative Normalization Preprocessing}
Forcing contextual embedding vectors in $X_s$ and $Y_s$ to be zero-mean is straightforward. Nevertheless, it is difficult to look for the universal mean vector of contextual embeddings when we train the English parser, because we do not have such an exact mean vector for all possible contextual embeddings. Thus, to successfully implement IN for pre-training the parser, we calculate the approximate universal mean vector by averaging all contextual vectors of every occurrence of tokens from the given training dataset in each iteration. IN runs for 5 iterations, which is sufficient for convergence. %, excluding vectors for special tokens (`[CLS]’, `[SEP]’ and `[PAD]’). This is also applied to the test dataset during evaluation. In our experiments, we set iteration number as 5.

\section{Discussion}
\subsection{Why Contextual Embedding Mapping?}
\paragraph{Compare with Previous Methods:} Overall results are shown in Table \ref{tab:results}. In the first place, our contextual-aware embedding mapping (row 4 - 7) exceeds the aligned fastText (row 1) by a large margin. Moreover, our sense-level mapping without IN preprocessing outperforms uncased and cased mBERT by 0.67\% and 1.42\% on LAS on average, and mapping with preprocessing further outperforms them by 2.07\% and 2.82\% on average.
% Although mBERT provides representations for 104 languages, quality of representations may not be created equally across languages \citep{wu-dredze-2020-languages}. Using bilingual space mapping is recommended when it comes to zero-shot transfer between only two languages.
%\textbf{to build a concept-shared semantic space}
%Precise representation is a critical key for cross-lingual mapping. However, representations for context-independent embeddings are hard to well fit cross-lingual spaces due to less precision, which is credibly indicated in Table \ref{tab:results} -- 
\paragraph{Dictionary-free Mapping:}
Typically, aligned embeddings take a static dictionary as reference but high-quality manual dictionaries are still rare \citep{ruder2019survey}. Our mapping skips the word-level alignment in dictionaries, and directly aligns the embeddings from parallel corpora which offers a large scope of token alignments.
\paragraph{Sense-level Mapping:}
Different from static embeddings whose words only have one unique representation, our contextual embeddings also take advantage of multiple representations for multi-sense words to improve the quality of anchor-driven mapping. In Table \ref{tab:results}, the performance of sense-level mapping always surpasses word-level mapping.

%, where it is impressive that anisotropic degree is dramatically reduced  only after first step

\subsection{Effect of Iterative Normalization}
Figure \ref{fig:horizon-chart} illustrates the various degrees of anisotropy among different language pairs. As we expect, the anisotropic degree for English (pink, right) is basically constant, but there is large discrepancy between other target languages (blue, left). After IN preprocessing, all language spaces are approximately isotropic, where their scores of anisotropy dramatically reduce near to zero. One example of how the anisotropic degree drops down in each iteration of IN for the Spanish-English pair is illustrated in Figure \ref{fig:IN-reduction}. IN assists the aligned embeddings in building more similar relative 
angles across embeddings in different language spaces. As shown in Table \ref{tab:results}, this preprocessing improves an absolute gain of 1.37\% for word-level mapping and  1.40\% for sense-level mapping on average.

\section{Conclusion}
We proposed a linear, dictionary-free and sense-level contextual mapping approach by exploiting parallel corpus which has shown promising results and substantial improvement compared with multilingual fastText and mBERT in the zero-shot dependency parsing task. We also revealed that various degrees of anisotropy hurts the performance of mapping, and introduced iterative normalization to alleviate it by enforcing contextual embeddings to be uniformly distributed, which also has indicated the benefits of isotropy.
\section*{Acknowledgments}
We thank the anonymous reviewers for their valuable comments.

% The acknowledgments should go immediately before the references. Do not number the acknowledgments section.
% Do not include this section when submitting your paper for review.

\bibliography{anthology,eacl2021}
\bibliographystyle{acl_natbib}

\clearpage
\appendix

\section{Pseudo Code of Contextual Embedding Transformation}
Pseudo Code is shown in Alogirthm \ref{alg:cet}.
\label{appendix:algo}
\begin{algorithm*}[h] 
\begin{small}
\caption{Contextual Embedding Transformation}
\label{alg:cet}
\begin{algorithmic}[1]
\Require{Target Corpus $\mathcal{X}$, source Corpus $\mathcal{Y}$, target pre-trained BERT $\mathcal{B}_x$, source pre-trained BERT $\mathcal{B}_y$, where $\mathcal{X}$ is the translation corpus of $\mathcal{Y}$}
\Statex
\Function{Contextual-Transformation}{$\mathcal{X}$, $\mathcal{Y}$, $\mathcal{B}_x$, $\mathcal{B}_y$}
    \State \# Part 1: Collect embeddings
    \State {$\mathcal{I}$ $\gets$ \Call{Fast-Align}{$\mathcal{X}$, $\mathcal{Y}$}} \Comment{$\mathcal{I}$ is an index-aligned corpus, where each line is composed of index pairs of aligned tokens for each parallel sentence.}
    \State Initialize $\mathcal{C} \gets$ Empty Hash Map
    \For{index i in \Call{Length}{$\mathcal{X}$}}    \Comment{number of sentences in the corpus}
        \State $X \gets \mathcal{X}[i]$, $Y \gets \mathcal{Y}[i]$,  $I \gets \mathcal{I}[i]$
        \State{$E_X \gets \mathcal{B}_x(X)$} \Comment{Contextual embeddings of tokens: }
        \State{$E_Y \gets \mathcal{B}_y(Y)$}
        \For{index j in \Call{Length}{X}} \Comment{number of tokens in the sentence}
            \State $x \gets X[j]$, $e_x \gets E_X[j]$
            \State $e_y \gets E_Y[I(j)]$ \Comment{Find the aligned embedding by looking at $I$, where $I(j)$ is the index of aligned source token.}
            \State $\mathcal{C}[x]$.append($(e_x, e_y)$)
        \EndFor
    \EndFor
    \State 
    \State \# Part 2: Obtain aligned sense-level embeddings
    \State Initialize Empty matrix $X_s$, $Y_s$
    \For{target type $x$ in $\mathcal{C}$.keys()}
        \State $c_x \gets$ \text{all target embeddings } $e_x$ in $\mathcal{C}[x]$
        \State $c_y \gets$ \text{all target embeddings } $e_y$ in $\mathcal{C}[x]$
        \State $k \gets$ \Call{Elbow-based}{$c_x$} \Comment{Find optimal number of clusters}
        \For{Subcluster $c_{x_i}$ in \Call{K-Means}{$k$, $c_x$}}
            \State Get subcluster $c_{y_i}$ due to aligned pair ($(e_x, e_y)$) in $\mathcal{C}[x]$
            \State $mean_x \gets$ mean vector of $c_{x_i}$
            \State $mean_y \gets$ mean vector of $c_{y_i}$
            \State Put $mean_x$ in $X_s$ as a column
            \State Put $mean_y$ in $Y_s$ as a column
        \EndFor
    \EndFor
    \State
    \State \# Part 3: Derive embedding transformation
    \State $U\Sigma V^T = \text{svd}(YX^T)$
    \State $\hat{W} = UV^T$
    \State \Return {$\hat{W}$}
            
\EndFunction
\end{algorithmic}
\end{small}
\end{algorithm*}

\section{Hyperparamters}
\label{sec:appendix:hyper}
Here we list all hyperparamters for our pre-trained parser in Table \ref{tab:hyper}.
\begin{table}[ht]
\begin{center}

  \begin{tabular}{cc}
    \hline
    Hyperparameters & Value \\
    \hline
     Batch size & 128 \\
     Arc representation dim & 500\\
     Tag representation dim & 100\\
     Dropout & 0.3 \\
     LSTM hidden size & 500 \\
     LSTM \# layers & 3 \\
     Pos tag embedding dim & 100 \\
     Grad norm & 5 \\
     \# epochs  & 200 \\
     Patience & 25 \\
     Optimizer & Dense Sparse ADAM\\
     Learning rate & 0.0008 \\
     Encoding Layer & Bi-LSTM\\
     
     \hline
    
  \end{tabular}
\end{center}
\caption{Hyperparameters for deep biaffine dependency parser training.}
\label{tab:hyper}
\end{table}

\section{pre-trained Monolingual BERTs}
\label{sec:appendix:BERT}
In Table \ref{tab:BERTs}, we list the names of pre-trained monolingual BERTs from huggingface \footnote{\url{https://huggingface.co/models}} that we used in our  experiments.

\begin{table}[ht]
\begin{center}
\begin{small}
  \begin{tabular}{cl}
    \hline
    Language & Model name \\
    \hline
    mbert uncased & bert-base-multilingual-uncased \\
    mbert cased & bert-base-multilingual-cased \\
    en & bert-base-uncased \\
    es & dccuchile/bert-base-spanish-wwm-uncased \\
    pt & neuralmind/bert-base-portuguese-cased \\
    ro & dumitrescustefan/bert-base-romanian-uncased-v1 \\
    pl & dkleczek/bert-base-polish-uncased-v1 \\
    fi & bert-base-finnish-uncased-v1 \\
    el & nlpaueb/bert-base-greek-uncased-v1 \\
    \hline
    
     \hline
    
  \end{tabular}
  \end{small}
\end{center}
\caption{Names of Pre-trained BERT models.}
\label{tab:BERTs}
\end{table}

\end{document}